\documentclass[letterpaper, 10 pt, conference]{ieeeconf}  

\IEEEoverridecommandlockouts                              

\usepackage{tabularx}
\usepackage{adjustbox}
\usepackage{booktabs}
\usepackage{amsmath}
\usepackage{amsfonts}
\usepackage{wrapfig}
\usepackage{graphicx}
\usepackage{amsmath} 
\usepackage{amssymb}  
\usepackage{bm}
\usepackage[ruled]{algorithm2e}
\usepackage{algpseudocode}
\usepackage{textcomp}
\usepackage{lipsum}
\usepackage[table]{xcolor}
\usepackage{array,colortbl}
\usepackage[noadjust]{cite}

\usepackage{xcolor}

\usepackage{array}
\newcolumntype{P}[1]{>{\centering\arraybackslash}p{#1}}

\newcolumntype{N}{>{\centering\arraybackslash}m{.5in}}
\newcolumntype{G}{>{\centering\arraybackslash}m{2in}}
\newcolumntype{C}[1]{>{\centering\let\newline\\\arraybackslash\hspace{0pt}}m{#1}}

\usepackage{pdfpages}

\usepackage{hyperref}
\hypersetup{
    colorlinks=true,
    linkcolor=black,
    filecolor=magenta,      
    urlcolor=cyan,
    citecolor=black,
}

\title{\LARGE \bf
\vspace{10pt}
Learning Object Compliance via Young's Modulus from Single Grasps using Camera-Based Tactile Sensors
\vspace{-10pt}
}
\author{
    \authorblockN{Michael Burgess$^1$, Jialiang Zhao$^1$, Laurence Willemet$^1$} 
        \authorblockA{$^1$Massachusetts Institute of Technology (MIT)\\
    {\tt\small mburgjr@csail.mit.edu, alanzhao@csail.mit.edu, lwilleme@csail.mit.edu} 
    } }

\begin{document}

\maketitle
\thispagestyle{empty}
\pagestyle{empty}

\begin{abstract}

    Compliance is a useful parametrization of tactile information that humans often utilize in manipulation tasks. It can be used to inform low-level contact-rich actions or characterize objects at a high-level. In robotic manipulation, existing approaches to estimate compliance have struggled to generalize across both object shape and material. Using camera-based tactile sensors, proprioception, and force measurements, we present a novel approach to estimate object compliance as Young's modulus $E$ from parallel grasps. We evaluate our method over a novel dataset of 285 common objects, including a wide array of shapes and materials with Young's moduli ranging from 5.0~kPa to 250~GPa. Combining analytical and data-driven approaches, we develop a hybrid system using a multi-tower neural network to analyze a sequence of tactile images from grasping. This system is shown to estimate the Young's modulus of unseen objects within an order of magnitude at 74.2\% accuracy across our dataset. This is an improvement over purely analytical and data-driven baselines which exhibit 28.9\% and 65.0\% accuracy respectively. Importantly, this estimation system performs irrespective of object geometry and demonstrates increased robustness across material types. Code is available on \href{https://github.com/GelSight-lab/TactileEstimateModulus/tree/main}{GitHub} and collected data is available on \href{https://huggingface.co/datasets/mburgjr/GelSightYoungsModulus}{HuggingFace}.

\end{abstract}

\section{Introduction}

    Humans naturally use tactile feeedback to characterize objects and inform our interactions with them~\cite{Navarro-Guerrero2023-td}. Compliance, or softness, is useful in everyday tasks. For example, we can determine the ripeness of an avocado or other fruits by assessing their compliance. Crucially, surgeons rely on tactile feedback and perceived compliance to accurately characterize tissues and detect tumors~\cite{hapticsurgical}. 

    In robotics, if we would like to better replicate the natural proficiency of human manipulation through teleoperated action, we must communicate this tactile information to remote users. Recently, haptic devices have been developed to display softness for remote users~\cite{mete2024sori, 7177690, 10693458}. However, these displays are only useful if we can reliably provide them a measure of compliance in real-time.
    
    \begin{figure}[htbp]
        \centering
        \includegraphics[width=\linewidth]{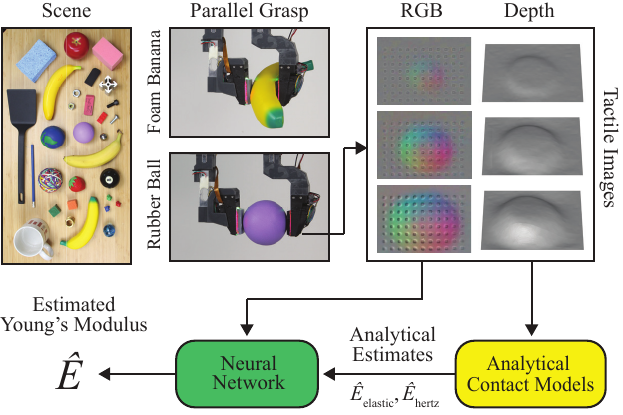}
        \vspace{-15pt}
        \caption{\footnotesize{\textbf{Young's modulus estimation system.} Tactile images from parallel grasps of various objects are fed into learned models alongside analytical estimates to generate a hybrid estimate of Young's modulus. Tactile images are size 35~mm by 25~mm.}}
        \label{fig:teaser}
        \vspace{-15pt}
    \end{figure}

    Humans sense discrepancies between soft and hard objects through fingertip pressure and deformation~\cite{Srinivasan1995-yy}. Camera-based tactile sensors, like GelSight, are specifically designed to track these parameters on robotic fingertips~\cite{LUO201754, yuan2017gelsight}. Analytical models, such as Hertzian mechanics, follow a similar approach to approximate compliance as humans may~\cite{johnson1987contact,popov2010contact}. Ideally, we could input tactile data to these models to determine compliance. However, traditional analytical frameworks often struggle to generalize across arbitrary geometries that may be encountered in unstructured environments. Since an object's deformation under load depends on both its shape and compliance, analytical methods cannot rigorously determine compliance without prior knowledge of object geometry~\cite{cutkosky1984mechanical}. 
    
    Alternatively, learning-based implementations have been developed to estimate compliance from tactile images alone without the use of mechanical modeling~\cite{yuan2017hardness, di2024using, fu2024touch, yu2024octopi}. While these methods have demonstrated success in estimating compliance independent of shape, they still face limitations in describing compliance across a wide range of materials. Many of these models elect to exclusively label the compliance of rubber objects via the Shore hardness scale. To successfully incorporate compliance feedback into a teleoperation interface, we need labels of compliance on a universal scale to replicate the continuity that humans experience in real-world manipulation.

    To address these limitations and move towards more generalized compliance estimation, we propose a hybrid approach that combines analytical modeling with learning-based methods to estimate an object's compliance as Young's modulus $E$ from a single parallel grasp. By incorporating analytical models, we improve generalization across materials, while using a learned CNN to account for geometric complexities. We label compliance as Young's modulus to align with analytical models and produce estimations on a universal scale.

    In this work, we present the following contributions:
    \begin{itemize}
        \item A novel compliance estimation architecture which demonstrates increased accuracy over the current state-of-the-art. Unlike previous methods, our hybrid learned architecture incorporates analytical modeling.
        \item Introduce Young's modulus $E$ as a label of compliance to enable seamless incorporation of analytical models and produce estimates on a universal scale.
        \item A novel dataset containing 285 common objects, featuring a wide variety of shapes and materials, which could be used in developing future compliance estimation architectures.
    \end{itemize}
    We validate that our hybrid estimation approach can generalize reasonably well across our diverse dataset. Thus, our estimation system could be applied in robotic manipulation systems to more robustly estimate compliance.


\section{Related Works}

    \subsection{GelSight Sensors \& Contact Modeling}

        Camera-based tactile sensors, such as GelSight, are designed to track contact geometry in high-resolution using a camera placed behind an elastomer gel pad~\cite{yuan2017gelsight,wang2021gelsight,tippur2024rainbowsight}. They are effective for sensing texture~\cite{Li_2013_CVPR}, object shape and pose~\cite{zhao2023fingerslam}, contact force~\cite{yuan2017gelsight}, and friction~\cite{7139016}. However, reliably modeling contact with these soft sensors remains challenging. Finite element method (FEM) has proven a viable option~\cite{8794113, taylor2022gelslim}. Although modeling with FEM is accurate, it requires a precise map of object geometry and solutions often must be pre-computed offline.

        Alternative approaches have been developed using Hertzian mechanics to model soft contact in robotic grasping~\cite{933086}. This closed-form method offers improved computational efficiency over FEM and other methods. Still, it requires assumptions on contact geometry, namely that geometry should be axisymmetric. Hertzian contact models have been utilized for force estimation~\cite{DBLP:conf/iros/McInroeCGBF18, di2024using} and curvature detection from tactile data~\cite{9197050}. These closed-form constitutive models can be inverted to determine compliance.

    \subsection{Compliance Estimation}

        Estimating the compliance of a contacted object remains a difficult problem in robotic manipulation, particularly without assumptions on object shape or material type. Analytical approaches to compliance estimation often lack robustness to variations in geometry. The most traditional approach to model compliance is to apply Hooke's Law to measurements of grasping force and position~\cite{cutkosky1984mechanical}. This method can provide an effective stiffness of grasping, but may struggle when applied to unknown objects with complex shapes. By incorporating tactile sensing, it is possible to create higher fidelity compliance estimation models~\cite{6943124, useOfTactileFeedback}. These methods have a stronger ability to generalize across contact geometry through sensing, but still struggle to account for complex contact geometries.

        Recent learning-based approaches to compliance estimation have demonstrated the ability to better generalize across geometry, but struggle when applied over a wide-range of material types. Data-driven architectures have been developed to estimate the Shore 00 hardness of an object in contact with a GelSight sensor directly from tactile images without any mechanical modeling~\cite{yuan2017hardness}. The Shore 00 hardness scale is designed to quantify the compliance of soft rubbers~\cite{Gent1958OnTR}. This model is trained using exclusively rubber objects and is thus limited to this material domain. Alternatively, LLM-backed methods have been developed to classify tactile images with binary semantic descriptions of ``soft'' or ``hard''~\cite{fu2024touch, yu2024octopi}. These large data models are promising, but have not been used to assess finer differences in compliance on a continuous or quantitative scale. Other data-driven implementations have been developed using tactile sensing to directly detect the ripeness of produce~\cite{He2023, ERUKAINURE2022107289, avocadomaturity}. As before, these methods are restricted by their dataset and labeling method. In all cases, models do not incorporate analytical modeling alongside learning.


\begin{figure*}[t!]
    \centering
        \includegraphics[width=0.875\linewidth, ext=.png, read=.pdf]{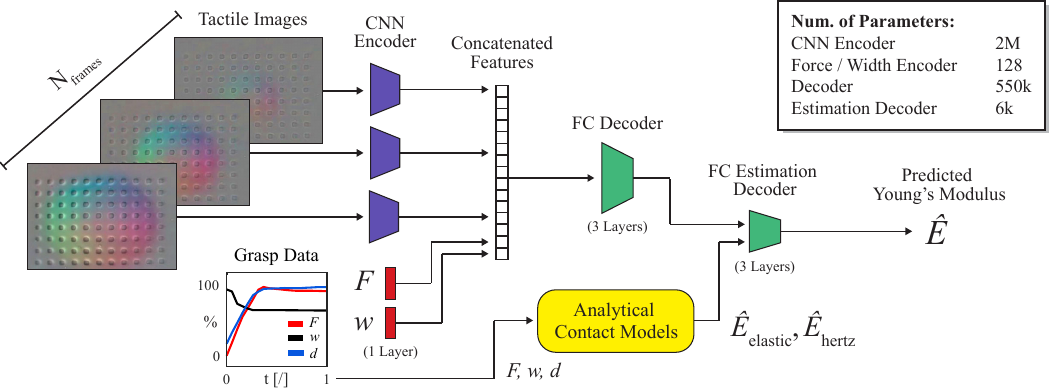}
        \vspace{-5pt}
    \caption{\footnotesize{\textbf{Young's modulus estimation architecture.} Tactile images, depth, and measured grasping force $F$ and width $w$ are sent into analytical and learned models. Estimates from analytical models are fed into a fully-connected decoder alongside learned grasp features to produce hybrid estimate $\hat{E}$.}}
    \label{fig:arch}
    \vspace{-15pt}
\end{figure*}

\section{Methods}

    \subsection{Overview}

        We develop a hybrid system that fuses analytical models alongside learning to estimate the Young's modulus of general unknown objects from a single parallel grasp. Through analytical modeling, we can create a well-founded, preliminary estimate for Young's modulus. By incorporating learning, we can compensate for assumptions taken by analytical models to better generalize across contact geometry. A large, diverse dataset of 285 common objects is collected to train and evaluate our method. This approach of combining analytical models with data-driven methods is partially inspired by residual physics architectures~\cite{zeng2019tossingbot} and physics-informed tactile models~\cite{huang2022understanding}.
        
        Our hybrid estimation architecture is shown in Fig.~\ref{fig:arch}. After grasping, tactile images and measurements of normal force $F$ and gripper width $w$ are input into analytical models to fit respective estimates of Young's modulus $\hat{E}_\text{elastic}, \hat{E}_\text{hertz}$. In parallel, a selected number $N_\text{frames}$ of tactile images from a single sensor are passed to a CNN. Our CNN architecture is adapted from \cite{kim2023simultaneoustactileestimationcontrol}, which has demonstrated strong performance in classifying sequences of tactile images. Features extracted from tactile images by this CNN are concatenated with grasp measurements $F$, $w$ and fed into a large fully-connected decoder. Output features from the large decoder are sent alongside analytical estimates to a smaller decoder to produce a final hybrid estimate of Young's modulus $\hat{E}$. This multi-tower network is inspired by other grasp-based architectures~\cite{mahler2017dexnet}.

        Code for the project is publicly available on \href{https://github.com/GelSight-lab/TactileEstimateModulus/tree/main}{GitHub}. Collected data with labeled Young's moduli and Shore hardnesses is publicly available on \href{https://huggingface.co/datasets/mburgjr/GelSightYoungsModulus}{HuggingFace}.

    \subsection{Analytical Contact Models}\label{sec:contactmodels}

\begin{figure*}[!h]
            \centering
                \includegraphics[width=\linewidth, ext=.png, read=.pdf]{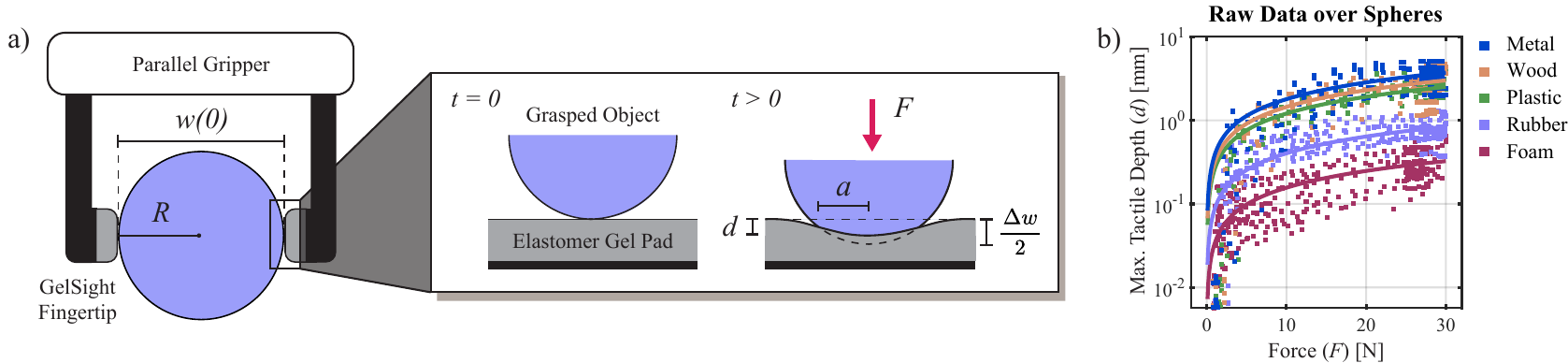}
                \vspace{-20pt}
            \caption{\footnotesize{\textbf{Parallel grasping and Hertzian mechanics}. (a) Sphere of radius $R$ is grasped with soft GelSight sensors. Normal force $F$ is applied at width $w$. Tactile penetration depth $d$ and contact radius $a$ are measured with tactile sensors. Timestep $t = 0$ is the moment of first contact over the grasp. (b) Normal force $F$ and maximum tactile depth $d$ measurements are collected across grasps for spheres of each material type and fit to $d \sim F^\frac{2}{3}$, per our Hertzian model.}}
            \label{fig:hertz_and_raw_data}
            \vspace{-15pt}
        \end{figure*}

        \textbf{\textit{Simple Elasticity:}} Parallel grasping can be modeled as uniaxial loading following Hooke's Law. A diagram of parallel grasping is provided in Fig.~\ref{fig:hertz_and_raw_data}a. At every instance $t$, normal force $F$ is applied to the grasped object, with gripper at width $w$. Traditional methods have used these measurements alone to estimate the stiffness of grasped objects~\cite{cutkosky1984mechanical}. Throughout this paper, we assume objects to be made of isotropic materials.

        With camera-based tactile sensors, we can increase the fidelity of contact models by measuring contact area $A$ and depth $d$~\cite{yuan2017gelsight}. Contact area $A$ is determined by masking tactile depth images with a constant threshold of 0.1mm. This threshold was chosen empirically to cancel out potential noise. Without tactile sensing, we could not measure depth ($d(t) = 0$) and would need to assume a constant contact area ($A(t) = A_\text{sensor}$). We calculate contact stress $\sigma$ and object strain $\epsilon$, following Eq.~\ref{eq:stress} and Eq.~\ref{eq:strain}. This model assumes symmetry between parallel fingers. We define $t = 0$ to be the moment of first contact, where $F(t = 0)$ must be greater than an impirically chosen threshold of 0.75~N.
        \begin{equation}
                \sigma(t) = \frac{F(t)}{A(t)}
                \label{eq:stress}
        \end{equation}
        \footnotesize\begin{equation}
                \epsilon(t) = \frac{\Delta w(t) + 2 \Delta d(t)}{w(0) + 2 d(0)} = \frac{(w(t) + 2d(t)) - (w(0) + 2 d(0))}{w(0) + 2 d(0)}
                \label{eq:strain}
        \end{equation}\normalsize
        Young's modulus is defined as the ratio between elastic stress and strain~\cite{hibbeler1994mechanics}. Using measurements across grasping time $t$, we can fit a linear estimate for Young's modulus $\hat{E}_\text{elastic}$ following Eq.~\ref{eq:simple_elasticity}.
        \begin{equation}
                \boldsymbol{\sigma} = \hat{E}_\text{elastic} \boldsymbol{\epsilon}
                \label{eq:simple_elasticity}
        \end{equation}
        Despite its simplicity, this method does not accurately calculate object stress or strain, especially if the object is larger than the sensor surface. Furthermore, it overlooks precise information about the deformation of contact geometry that is apparent through high-resolution tactile data.

        \textbf{\textit{Hertzian Contact:}} Hertzian contact theory models the deformation of elastic bodies in contact with each other. In a simple case, contact is assumed to be without friction or adhesion~\cite{popov2010contact, johnson1987contact}. Unlike previous methods, Hertzian mechanics closely analyzes geometry through contact. It has been applied to model soft contact in robotics with camera-based tactile sensors~\cite{933086, 9197050, taylor2022gelslim, di2024using}.

        GelSight sensors are made with elastic silicone rubber~\cite{wang2021gelsight}. Unless a grasped object is rigid, we expect both bodies to deform through contact. This implies the maximum measured tactile depth $d$ will be less than the relative displacement between the finger and object $d < \frac{\Delta w}{2}$, as depicted in Fig.~\ref{fig:hertz_and_raw_data}a. An aggregate modulus $E^*$ describes the bodies' combined resistance to deformation in Eq.~\ref{eq:E_star}, which depends on the Poisson's ratio $\nu$ of each body.    
        \begin{equation}
                E^* = \left(\frac{1 - \nu_\text{sensor}^2}{E_\text{sensor}} + \frac{1 - \nu_\text{obj}^2}{E_\text{obj}}\right)^{-1}
        \label{eq:E_star}
        \end{equation}
        Hertzian solutions have been derived for a wide range of contact geometries~\cite{popov2019handbook}. We modeled the sensor as a half-space, which may be a reasonable approximation in most cases but could be less accurate for large, rigid objects that penetrate deeper into the sensor's finite thickness. We modeled all grasped objects as spherical. This assumption is reasonable for convex objects, but may produce bounded errors over our geometrically diverse dataset. We expect that out neural network can learn to compensate for this intrinsic and repeatable geometric modeling error.

        Since we assume contact profiles are spherical, we calculate contact radius $a$ based on observed total contact area $A$. Additionally, we calculate the apparent radius of the unknown object $R$ using contact area and gripper displacement~\cite{popov2019handbook}. Using these parameters, we can relate tactile depth $d$ to normal force $F$ as formulated in Eq.~\ref{eq:MDR}. This relationship is derived via the method of dimensionality reduction (MDR) in Appendix~\ref{appendix:MDR}~\cite{johnson1987contact, popov2019handbook, MDRarticle}. With equations established and measurements over the duration of a grasp, we can apply least-squares to retrieve a best fit value $\hat{E}^{*}$.
        \begin{equation}
                a(t) = \sqrt{\frac{A(t)}{\pi}} , \quad R(t) = \frac{a^2(t)}{\Delta w(t)}
                \label{eq:a_and_R}
        \end{equation}
        \begin{equation}                
                \boldsymbol{d} = \frac{1 - \nu_\text{sensor}^2}{E_\text{sensor}} {\left(\frac{3 {{\hat{E}}^{* 2}} \boldsymbol{F}}{32 \boldsymbol{R}^2}\right)}^{\frac{1}{3}} \boldsymbol{a}
                \label{eq:MDR}
        \end{equation}
        With calculated $\hat{E}^{*}$, we can extract an estimated modulus for an unknown object $\hat{E}_\text{hertz}$ through Eq.~\ref{eq:E_relation}, given known properties of our sensor $E_\text{sensor} = 0.275$ MPa, $\nu_\text{sensor} = 0.48$. We assume constant $\nu_\text{obj} = 0.4$ based on the Poisson's ratio of common materials~\cite{Greaves2011-tp}. After estimates are computed, a linear scaling is added to account for potential constant errors across the dataset.
        \begin{equation}
            \hat{E}_\text{hertz} = \left(1 - \nu_\text{obj}^2\right) {\left(\frac{1}{\hat{E}^*} - \frac{1 - \nu_\text{sensor}^2}{E_\text{sensor}}\right)}^{-1}
            \label{eq:E_relation}
        \end{equation}


\section{Physical Dataset}
        
\begin{figure}[htbp]
        \vspace{-5pt}
        \centering
         \includegraphics[width=\linewidth]{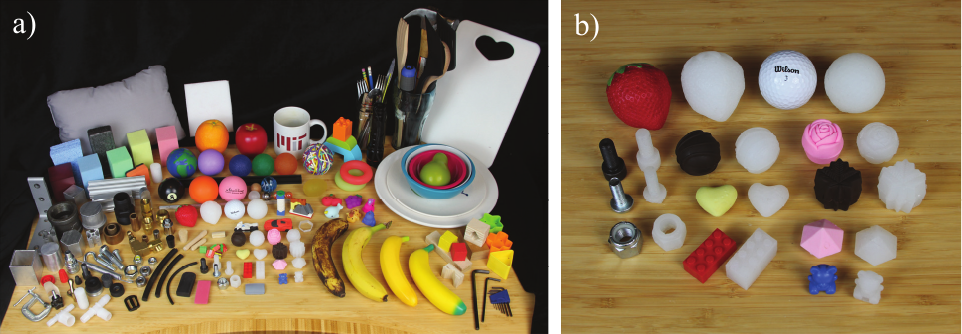}
         \vspace{-20pt}
        \caption{\footnotesize{\textbf{Physical dataset.} (a) Selected objects from our physical dataset are displayed. (b) Custom objects are molded out of silicone to replicate the shape of non-rubber objects in the dataset.}}
        \label{fig:dataset}
        \vspace{-7.5pt}
    \end{figure}

    A novel physical dataset of 285 common household objects was gathered to train and validate methods. For comparison, this is nearly 4 times larger than the YCB dataset, which includes only 77 objects~\cite{Calli2017-sw}. Additionally, each object was required to have approximately uniform material composition to ensure that an accurate Young's modulus could be defined for the object and used to label data for supervised training. Furthermore, chosen objects for training must not exhibit viscoelastic behavior. A subset of total dataset objects is displayed in Fig.~\ref{fig:dataset}. Some objects, such as bananas and other real fruit, are not used for model training.

    To evaluate estimation algorithms' ability to generalize, diversity of shape and material were prioritized in acquiring objects. We mold 20 custom objects out of silicone to replicate the exact shape of objects in the dataset that are made of much harder materials. Molded objects are displayed in Fig.~\ref{fig:dataset}b. Training on data collected across object replicas of different materials may help prevent overfitting to shapes and textures. The dataset's composition is detailed in Tables~\ref{table:dataset_mat} and \ref{table:dataset_shapes}.

    \begin{table}[h]
        \small
        \vspace{2pt}
        \begin{minipage}{.45\linewidth}
        \caption[font=small,labelfont=bf]{Dataset Materials}\label{table:dataset_mat}
        \vspace{-8pt}
        \centering
                \begin{tabular}{cc}
                \toprule
                \textbf{Material} & {\textbf{Percent (\%)}} \\
                \midrule
                Rubber & 28.6\% \\
                Metal & 25.1\% \\
                Plastic & 24.0\% \\
                Wood & 8.4\% \\
                Foam & 6.3\% \\
                Other & 7.6\% \\
                \bottomrule
                \end{tabular}
        \end{minipage}
        \hfill
        \begin{minipage}{.45\linewidth}
        \caption[font=small,labelfont=bf]{Dataset Shapes}\label{table:dataset_shapes}
        \vspace{-8pt}
        \centering
                \begin{tabular}{cc}
                \toprule
                \textbf{Shape} & {\textbf{Percent (\%)}} \\
                \midrule
                Cylinder & 20.2\% \\
                Sphere & 11.5\% \\
                Rectangular & 15.3\% \\
                Hexagonal & 2.1\% \\
                Irregular & 50.9\% \\
                \bottomrule
                \end{tabular}
        \end{minipage}
        \vspace{-15pt}
    \end{table}

    As objects were gathered, they were labeled with their respective Young's moduli. Hard objects, made of plastics, wood, glass, and metals, were labeled by referencing an engineering database of known material properties~\cite{MatWeb}. This was possible given that all selected hard objects are made of known materials. While variations due to manufacturing conditions or other factors are possible, these labels are expected to be relatively accurate with respect to our vast material domain. 
    
\begin{figure}[htbp]
        \vspace{-5pt}
        \centering
         \includegraphics[width=\linewidth]{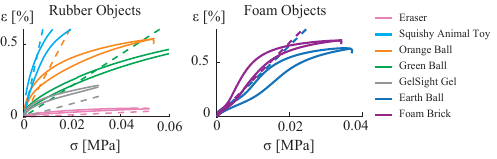}
         \vspace{-20pt}
        \caption{\footnotesize{\textbf{Dynamic Mechanical Analyzer results.} Compressive stress-strain curves recorded on samples of selected foam and rubber objects from our dataset. Dotted lines respresent expected linear curves from labeled Young's modulus $E$ for each object, where moduli were approximated via Shore hardness.}}
        \label{fig:dma}
        \vspace{-7.5pt}
    \end{figure}

    Soft objects, made of rubber and foam, were labeled by converting Shore hardness measurements to Young's modulus using Gent's model and other conversion methods~\cite{Gent1958OnTR, 10.5254/1.3547752, siliconekent}. Each soft object was indented with a durometer 10 times and the median measurement was taken. Conversion models provide a reliable measure of object compliance on a logarithmic scale, making them a widely used approximation in material characterization~\cite{mete2024sori}. To further validate assigned mechanical properties, 7 rubber and foam objects, including our sensor's gel pad, were probed with a Dynamic Mechanical Analyzer (TA Instrument Q850) over a force range of 0 to 10~N at a rate of 2~N/min. Results from this testing are plotted in Fig.~\ref{fig:dma} confirming the accuracy of our labeling techniques. In total, the dataset includes materials ranging from foam to steel, with Young's moduli from 5.0~kPa to 250~GPa. For reference, our sensor's Young's modulus is 0.275~MPa.
    
\begin{figure*}[!t]
    \centering
    \includegraphics[width=\linewidth]{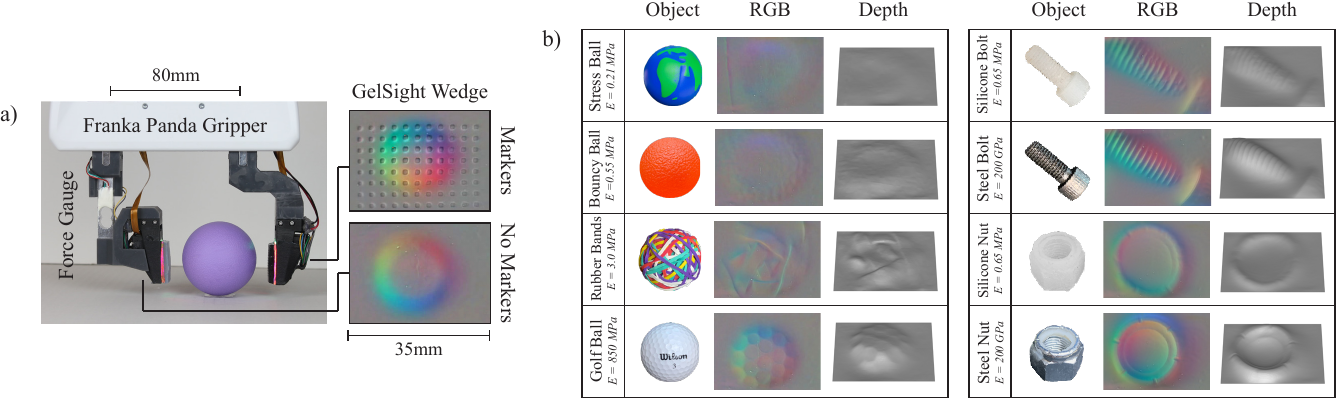}
    \vspace{-15pt}
    \caption{\footnotesize{\textbf{Hardware for data collection and example tactile data.} (a) GelSight Wedge sensors are attached to Franka Panda gripper with embedded force gauge. (b) Tactile images and depth reconstruction from a GelSight Wedge sensor without markers over a set of example objects. Shown frames are all from peak pressing force of 30~N. All displayed tactile images are size 35mm by 25mm.}}
    \label{fig:example_data_and_hardware}
    \vspace{-15pt}
\end{figure*}

\section{Experiments}

    \subsection{Hardware Setup}\label{sec:hardware}
        
        GelSight Wedge sensors~\cite{wang2021gelsight} were rigidly attached to a Franka Panda parallel robotic gripper with 3D-printed fingers, as depicted in Fig.~\ref{fig:example_data_and_hardware}a. Fingers were printed using FormLabs Tough 2000 resin. Although we used GelSight sensors for our experiment, other camera-based tactile sensors capable of tracking contact depth in high-resolution would be suitable. One sensor was inscribed with black dot markers to help enhance the dynamic information in a tactile image by tracking surface displacement~\cite{yuan2017gelsight,wang2021gelsight}. Data from only one sensor is required for estimation, but we collected data from both sensors simultaneously to later assess the necessity of markers in estimating Young's modulus. Additionally, we assume symmetry between fingers in our analytical models. 
        
        Normal force $F$ was measured through an HX711 force gauge in the left finger. Gripper position was obtained directly from control software. To improve accuracy, measured force was used to correct gripper position readings by compensating for finger deflection with a fitted linear model.

    \subsection{Data Collection \& Training}\label{sec:training}

        \textbf{\textit{Data Collection:}} Automated parallel grasps were executed on objects in fixed position. For each object in our dataset, a set number of grasps were recorded. Objects were re-oriented between each grasp to diversify potential antipodal grasp locations. During grasping, tactile images from GelSight sensors, normal forces, and gripper widths, were recorded at 30~Hz. Measurements were synchronized by shifting for latency between cameras. Grasping was executed at constant velocity of 3.75~cm/s and held at closed position for 0.5~s. Peak normal force varied across each grasp, up to 30~N. Raw data collected from grasps of spherical objects is plotted in Fig.~\ref{fig:hertz_and_raw_data}b.

        Tactile and depth images at grasping force of 30~N for a set of example objects are depicted in Fig.~\ref{fig:example_data_and_hardware}b. As expected from Eq.~\ref{eq:MDR}, peak depth is higher for objects of the same shape with higher modulus. Additionally, textures are more finely captured for rigid objects with higher modulus.

        \textbf{\textit{Data Preprocessing:}} Data recorded from grasping objects was clipped to include only the loading sequence, spanning from initial contact to peak force. Misaligned grasps where loading is non-monotonic were disregarded. A fixed number of frames ($N_\text{frames} = 3$) were sampled equidistantly in time across the loading sequence. To enhance training diversity, multiple sets of these samples were created from each grasp by shifting the starting point by one timestep. As illustrated in Fig.~\ref{fig:arch}, the sampled tactile images, along with analytical estimates, normal force $F$, and width $w$ measurements, were sent as inputs to our multi-tower architecture. Additionally, the images undergo random horizontal and vertical flipping for data augmentation.
        
        \textbf{\textit{Training:}} We trained our model for 80 episodes over 4,000 input grasps. In total, these grasps include over 12,000 tactile images. Training was performed end-to-end without any pretraining. Output labels were normalized on a log10 scale given the range of Young's moduli in our dataset. Final sigmoid activation function was used with an MSE loss function.


\section{Results}

\begin{figure}[!h]
        \centering
        \vspace{-15pt}
        \includegraphics[width=0.6\linewidth, ext=.png, read=.pdf]{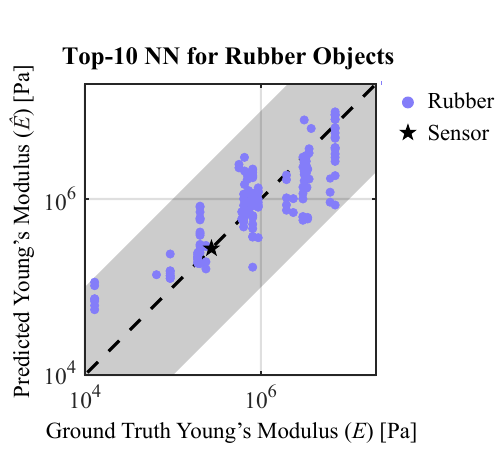}
        \vspace{-12.5pt}
        \caption{\footnotesize{\textbf{Young's modulus predictions for rubber objects only.} Results for model trained on only rubber objects over 80 randomly sampled grasps of unseen objects. Predictions in the gray region are considered sufficiently accurate within an order of magnitude of ground truth.}}
        \label{fig:rubber_only}
        \vspace{-5pt}
    \end{figure}

    Initially, we train an instance of our model using only rubber objects from our dataset. From Eq.~\ref{eq:E_relation}, it is expected to be easier to discern fine differences in compliance between rubber objects, as their Young's modulus is closest to that of our sensor. Estimations across unseen objects for this model are plotted in Fig.~\ref{fig:rubber_only}. The model demonstrates 100\% prediction accuracy within an order of magnitude of ground truth Young's modulus, with an average log10 error of 0.36. 
    
    \begin{figure*}[!t]
        \centering
        \includegraphics[width=\linewidth, ext=.png, read=.pdf]{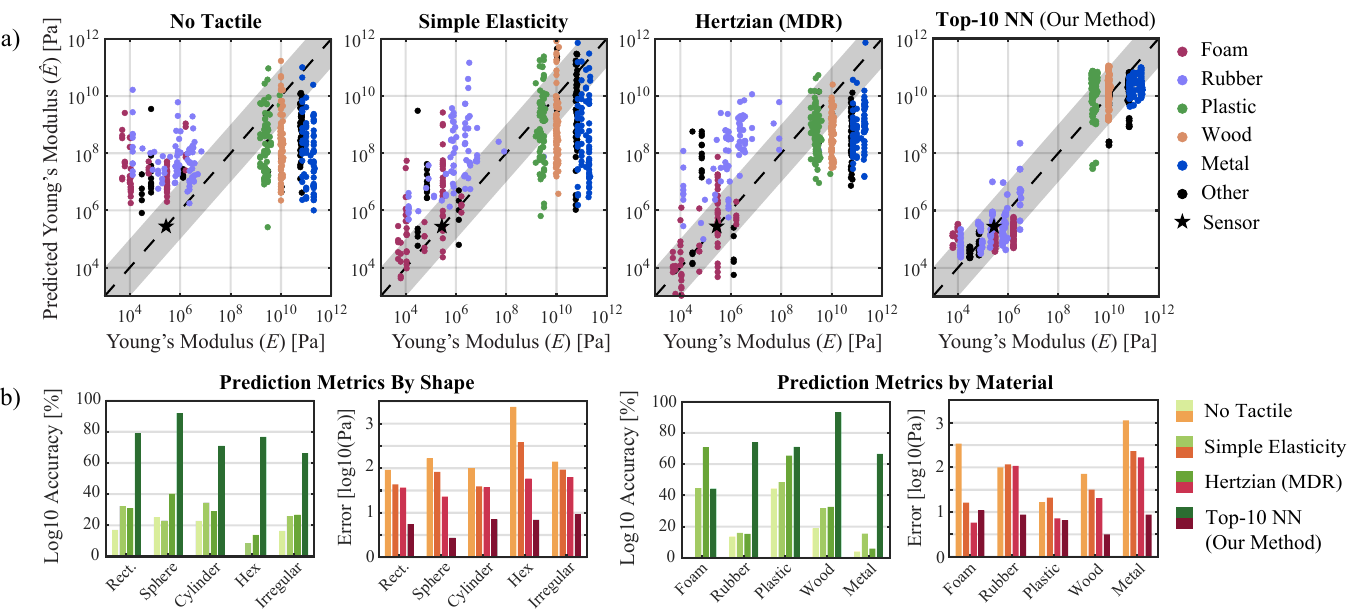}
        \vspace{-17.5pt}
        \caption{\footnotesize{\textbf{Young's modulus predictions and model performance.} (a) Results for learned and analytical estimation methods are plotted, using predictions from 80 randomly sampled grasps per material category. Predictions in the gray region are considered sufficiently accurate. (b) Log10 error and accuracy are plotted across dataset categories of shape and material. Neural network results are from top-10 trained models evaluated on unseen objects.}}
        \label{fig:predictions}
    \end{figure*}
    
    \begin{table*}[!tb]
        \footnotesize
        \centering
        \renewcommand{\arraystretch}{0.7}
        \setlength{\tabcolsep}{2pt}
        \vspace{-2.5pt}
        \begin{tabular}{C{2cm}C{1.8cm}C{1.25cm}C{1cm}C{1cm}C{1cm}C{1.8cm}C{1.8cm}C{1.8cm}C{1.8cm}}
            \toprule
            \multicolumn{1}{N }{} & \multicolumn{5}{c }{\textbf{Inputs}} & \multicolumn{2}{ c }{\textbf{Seen Objects}} & \multicolumn{2}{c }{\textbf{Unseen Objects}} \\
            \cmidrule(lr){2-6}
            \cmidrule(lr){7-8}
            \cmidrule(lr){9-10}
            \multicolumn{1}{ c }{\textbf{Method}} & Depth / RGB & Markers & $F$ & $w$ & \multicolumn{1}{ c }{$\hat{E}_\text{elastic}, \hat{E}_\text{hertz}$} & Log10 Error & Log10 Acc. & Log10 Error & Log10 Acc.  \\ 
            \cmidrule(lr){1-1}
            \cmidrule(lr){2-6}
            \cmidrule(lr){7-8}
            \cmidrule(lr){9-10}
            \multicolumn{1}{ c |}{No Tactile} & --- & --- & \checkmark & \checkmark & \multicolumn{1}{ c |}{---} & --- & \multicolumn{1}{ c |}{---} & 2.13 & 18.2\% \\ 
            \multicolumn{1}{ c |}{\quad Simple Elasticity \quad} & Depth & --- & \checkmark & \checkmark & \multicolumn{1}{ c |}{---} & --- & \multicolumn{1}{ c |}{---} & 1.90 & 27.3\% \\ 
            \multicolumn{1}{ c |}{Hertzian (MDR)} & Depth & --- & \checkmark & \checkmark & \multicolumn{1}{ c |}{---} & --- & \multicolumn{1}{ c |}{---} & 1.68 & 28.9\% \\ 
            \cmidrule(lr){1-1}
            \cmidrule(lr){2-6}
            \cmidrule(lr){7-8}
            \cmidrule(lr){9-10}
            \multicolumn{1}{ c |}{} & RGB & \checkmark & \checkmark & \checkmark & \multicolumn{1}{ c |}{\checkmark} & 0.73 & \multicolumn{1}{ c |}{76.9\%} & \textbf{0.76} & 73.5\% \\
            \multicolumn{1}{ c |}{} & RGB & \checkmark & $\times$ & $\times$ & \multicolumn{1}{ c |}{\checkmark} & 0.73 & \multicolumn{1}{ c |}{78.5\%} & \textbf{0.76} & \textbf{74.2\%} \\
            \multicolumn{1}{ c |}{} & RGB & $\times$ & \checkmark & \checkmark & \multicolumn{1}{ c |}{\checkmark} & 0.85 & \multicolumn{1}{ c |}{70.4\%} & 0.88 & 68.6\% \\
            \multicolumn{1}{ c |}{Top-10 NN} & RGB & $\times$ & $\times$ & $\times$ & \multicolumn{1}{ c |}{\checkmark} & 0.85 & \multicolumn{1}{ c |}{69.6\%} & 0.90 & 67.9\% \\
            \multicolumn{1}{ c |}{(Our Method)} & RGB & \checkmark & \checkmark & \checkmark & \multicolumn{1}{ c |}{$\times$} & 0.63 & \multicolumn{1}{ c |}{80.4\%} & 0.86 & 64.6\% \\
            \multicolumn{1}{ c |}{} & RGB & \checkmark & $\times$ & $\times$ & \multicolumn{1}{ c |}{$\times$} & 0.65 & \multicolumn{1}{ c |}{78.5\%} & 0.85 & 65.0\% \\
            \multicolumn{1}{ c |}{} & Depth & --- & \checkmark & \checkmark & \multicolumn{1}{ c |}{\checkmark} & 1.05 & \multicolumn{1}{ c |}{61.7\%} & 0.98 & 63.6\% \\
            \multicolumn{1}{ c |}{} & Depth & --- & $\times$ & $\times$ & \multicolumn{1}{ c |}{\checkmark} & 1.08 & \multicolumn{1}{ c |}{60.1\%} & 1.04 & 59.4\% \\
            \bottomrule
        \end{tabular}
        \caption{Estimation Architecture Ablation Study}
        \label{table:ablation}
        \vspace{-27.5pt}
    \end{table*}

    Furthermore, we utilize this reduced-domain training configuration to evaluate our architecture against previously-developed Shore hardness estimation models. We can only compare between predictions in this domain because previously-developed models have only been trained for use with rubber objects. Converting our model's predictions from Young's modulus to Shore 00 hardness~\cite{siliconekent}, we find that our model reduces root mean square error (RMSE) compared to Yuan et al.~\cite{yuan2017hardness} from 10.3 to 5.9 over simple-shaped objects and from 18.2 to 14.3 over arbitrarily-shaped objects, in units of Shore hardness. This observed improvement is likely due to our method's integration of analytical estimates alongside tactile image-based CNN models, whereas previous methods were fully data-driven.

    Subsequently, we trained our neural network over the entire collected dataset and compare against analytical baselines. Predictions across unseen objects for each method are plotted in Fig.~\ref{fig:predictions}a. A traditional Hooke's Law method without tactile sensing fails to distinguish between material types. Simple elastic and Hertzian models exhibit stronger performance, but remain less than 30\% accurate. Our hybrid learned model showcases dramatic improvements over all methods. Errors across all models may be partially attributed to potential inaccuracies in our dataset labelling process.

    In Table~\ref{table:ablation}, performance is compared between different ablations of our estimation methods and inputs. Performance metrics were computed for grasps of both seen and unseen objects. We evaluated models on a logarithmic scale given the range of Young's moduli in our dataset. We computed error as the average difference between prediction and ground truth on a log10 scale. Accuracy was computed based on the percentage of predictions within the same order of magnitude of ground truth labels.

    From the ablation study, we observe that our hybrid model out-performs purely analytical and learning-based methods. RGB tactile images with markers are shown to be most favorable for this estimation task. There is only minor differences in results between models which consider normal force $F$ and width $w$. This may be due to the fact that all objects were grasped with identical peak force. Moreover, force is implicitly apparent through tactile difference images~\cite{yuan2017gelsight}. 
    
    Importantly, consideration of analytical estimates $\hat{E}_\text{elastic}, \hat{E}_\text{hertz}$ is shown to improve prediction accuracy by nearly 10\%. This indicates our hybrid approach to compliance estimation is favorable over purely learning-based methods, which have been previously explored~\cite{yuan2017hardness, fu2024touch, yu2024octopi, He2023, ERUKAINURE2022107289, avocadomaturity}. As shown in Fig.~\ref{fig:predictions}b, our model performs independent of shape and can generalize well across materials. 

    \textbf{\textit{Limitations:}} Our estimation system struggled to express precise differences in Young's modulus between harder objects that stray further from the modulus of our sensor (i.e. plastic, wood, and metal objects). These results are expected given that harder objects are rigid relative to our soft sensor. Moreover, the resolution of our tactile sensor is not high enough to detect minute differences in deformation between this range of compliances. Future experiments could be performed with harder sensor gels to verify if this corresponds in an increased ability to sense the compliance of harder objects. Additionally, alternative analytical models that preport to better model the non-linear, volume-conserving behavior of silicone could be explored. However, this limitation may represent an intrinsic bottleneck in the precision of depth-tracking in camera-based tactile sensors.


\section{Conclusion}\label{sec:conclusion}

    We have developed a novel system for estimating the Young's modulus of unknown contacted objects using analytical and data-driven methods. This system can create online estimates from only a single parallel grasp using camera-based tactile sensors. It has been shown to perform independent of object shape and remains robust across material types. It can precisely distinguish between compliances of soft objects and is more accurate over this material domain than the current state-of-the-art~\cite{yuan2017hardness}. Moreover, by parametrizing compliance via Young's modulus, we are able to produce estimates on an absolute scale over many material categories. This generality is demonstrated through our diverse physical dataset.

    Given an ability to reliably produce estimations of compliance on a universal scale across different material regimes, this estimation system could be used alongside haptic displays to enhance teleoperation interfaces~\cite{mete2024sori}. With better tactile feedback to users, more performative policies for general robotic tasks and remote surgical operation could be developed. The estimation system has already been successfully deployed to measure compliance in real-time~\cite{willemet2025physicsinformed}.




\section*{ACKNOWLEDGMENTS}

    The authors would like to thank Megha Tippur, Sandra Q. Liu, and Yuxiang Ma for their valuable insights and guidance.


\bibliographystyle{IEEEtran}
\bibliography{refs}

\begin{thebibliography}{10}
\providecommand{\url}[1]{#1}
\csname url@samestyle\endcsname
\providecommand{\newblock}{\relax}
\providecommand{\bibinfo}[2]{#2}
\providecommand{\BIBentrySTDinterwordspacing}{\spaceskip=0pt\relax}
\providecommand{\BIBentryALTinterwordstretchfactor}{4}
\providecommand{\BIBentryALTinterwordspacing}{\spaceskip=\fontdimen2\font plus
\BIBentryALTinterwordstretchfactor\fontdimen3\font minus \fontdimen4\font\relax}
\providecommand{\BIBforeignlanguage}[2]{{%
\expandafter\ifx\csname l@#1\endcsname\relax
\typeout{** WARNING: IEEEtran.bst: No hyphenation pattern has been}%
\typeout{** loaded for the language `#1'. Using the pattern for}%
\typeout{** the default language instead.}%
\else
\language=\csname l@#1\endcsname
\fi
#2}}
\providecommand{\BIBdecl}{\relax}
\BIBdecl

\bibitem{Navarro-Guerrero2023-td}
N.~Navarro-Guerrero, S.~Toprak, J.~Josifovski, and L.~Jamone, ``\BIBforeignlanguage{en}{Visuo-haptic object perception for robots: an overview},'' \emph{\BIBforeignlanguage{en}{Auton. Robots}}, vol.~47, no.~4, pp. 377--403, Apr. 2023.

\bibitem{hapticsurgical}
M.~Zhou, D.~Jones, S.~Schwaitzberg, and C.~Cao, ``Role of haptic feedback and cognitive load in surgical skill acquisition,'' \emph{Proceedings of the Human Factors and Ergonomics Society Annual Meeting}, vol.~51, no.~11, pp. 631--635, 2007.

\bibitem{mete2024sori}
M.~Mete, H.~Jeong, W.~D. Wang, and J.~Paik, ``Sori: A softness-rendering interface to unravel the nature of softness perception,'' \emph{Proceedings of the National Academy of Sciences}, vol. 121, no.~13, 2024.

\bibitem{7177690}
M.~Bianchi, M.~Poggiani, A.~Serio, and A.~Bicchi, ``A novel tactile display for softness and texture rendering in tele-operation tasks,'' in \emph{2015 IEEE World Haptics Conference (WHC)}, 2015, pp. 49--56.

\bibitem{10693458}
Y.~Sun, F.~Meng, D.~Yang, M.~Xiong, and X.~Xu, ``Low-cost modeling and haptic rendering for membrane-like deformable objects in model-mediated teleoperation,'' \emph{IEEE Access}, vol.~12, pp. 141\,198--141\,210, 2024.

\bibitem{Srinivasan1995-yy}
M.~A. Srinivasan and R.~H. LaMotte, ``\BIBforeignlanguage{en}{Tactual discrimination of softness},'' \emph{\BIBforeignlanguage{en}{J. Neurophysiol.}}, vol.~73, no.~1, pp. 88--101, Jan. 1995.

\bibitem{LUO201754}
S.~Luo, J.~Bimbo, R.~Dahiya, and H.~Liu, ``Robotic tactile perception of object properties: A review,'' \emph{Mechatronics}, vol.~48, pp. 54--67, 2017.

\bibitem{yuan2017gelsight}
W.~Yuan, S.~Dong, and E.~H. Adelson, ``Gelsight: High-resolution robot tactile sensors for estimating geometry and force,'' \emph{Sensors}, vol.~17, no.~12, 2017.

\bibitem{johnson1987contact}
K.~L. Johnson, \emph{Contact mechanics}.\hskip 1em plus 0.5em minus 0.4em\relax Cambridge university press, 1987.

\bibitem{popov2010contact}
V.~L. Popov \emph{et~al.}, \emph{Contact mechanics and friction}.\hskip 1em plus 0.5em minus 0.4em\relax Springer, 2010.

\bibitem{cutkosky1984mechanical}
M.~R. Cutkosky, \emph{Mechanical properties for the grasp of a robotic hand}.\hskip 1em plus 0.5em minus 0.4em\relax Department of Computer Science, Carnegie-Mellon University, 1984.

\bibitem{yuan2017hardness}
W.~Yuan, C.~Zhu, A.~Owens, M.~A. Srinivasan, and E.~H. Adelson, ``Shape-independent hardness estimation using deep learning and a gelsight tactile sensor,'' in \emph{2017 IEEE International Conference on Robotics and Automation (ICRA)}.\hskip 1em plus 0.5em minus 0.4em\relax IEEE, May 2017.

\bibitem{di2024using}
J.~Di, Z.~Dugonjic, W.~Fu, T.~Wu, R.~Mercado, K.~Sawyer, V.~R. Most, G.~Kammerer, S.~Speidel, R.~E. Fan, G.~Sonn, M.~R. Cutkosky, M.~Lambeta, and R.~Calandra, ``Using fiber optic bundles to miniaturize vision-based tactile sensors,'' 2024.

\bibitem{fu2024touch}
L.~Fu, G.~Datta, H.~Huang, W.~C.-H. Panitch, J.~Drake, J.~Ortiz, M.~Mukadam, M.~Lambeta, R.~Calandra, and K.~Goldberg, ``A touch, vision, and language dataset for multimodal alignment,'' 2024.

\bibitem{yu2024octopi}
S.~Yu, K.~Lin, A.~Xiao, J.~Duan, and H.~Soh, ``Octopi: Object property reasoning with large tactile-language models,'' 2024.

\bibitem{wang2021gelsight}
S.~Wang, Y.~She, B.~Romero, and E.~Adelson, ``Gelsight wedge: Measuring high-resolution 3d contact geometry with a compact robot finger,'' in \emph{2021 IEEE International Conference on Robotics and Automation (ICRA)}.\hskip 1em plus 0.5em minus 0.4em\relax IEEE, 2021.

\bibitem{tippur2024rainbowsight}
M.~H. Tippur and E.~H. Adelson, ``Rainbowsight: A family of generalizable, curved, camera-based tactile sensors for shape reconstruction,'' in \emph{2024 IEEE International Conference on Robotics and Automation (ICRA)}.\hskip 1em plus 0.5em minus 0.4em\relax IEEE, 2024, pp. 1114--1120.

\bibitem{Li_2013_CVPR}
R.~Li and E.~H. Adelson, ``Sensing and recognizing surface textures using a gelsight sensor,'' in \emph{Proceedings of the IEEE Conference on Computer Vision and Pattern Recognition (CVPR)}, June 2013.

\bibitem{zhao2023fingerslam}
J.~Zhao, M.~Bauza, and E.~H. Adelson, ``Fingerslam: Closed-loop unknown object localization and reconstruction from visuo-tactile feedback,'' 2023.

\bibitem{7139016}
W.~Yuan, R.~Li, M.~A. Srinivasan, and E.~H. Adelson, ``Measurement of shear and slip with a gelsight tactile sensor,'' in \emph{2015 IEEE International Conference on Robotics and Automation (ICRA)}, 2015.

\bibitem{8794113}
D.~Ma, E.~Donlon, S.~Dong, and A.~Rodriguez, ``Dense tactile force estimation using gelslim and inverse fem,'' in \emph{2019 International Conference on Robotics and Automation (ICRA)}, 2019.

\bibitem{taylor2022gelslim}
I.~H. Taylor, S.~Dong, and A.~Rodriguez, ``Gelslim 3.0: High-resolution measurement of shape, force and slip in a compact tactile-sensing finger,'' in \emph{2022 International Conference on Robotics and Automation (ICRA)}.\hskip 1em plus 0.5em minus 0.4em\relax IEEE, 2022.

\bibitem{933086}
Y.~Li and I.~Kao, ``A review of modeling of soft-contact fingers and stiffness control for dextrous manipulation in robotics,'' in \emph{Proceedings 2001 ICRA. IEEE International Conference on Robotics and Automation (Cat. No.01CH37164)}, vol.~3, 2001.

\bibitem{DBLP:conf/iros/McInroeCGBF18}
B.~W. McInroe, C.~L. Chen, K.~Y. Goldberg, R.~Bajcsy, and R.~S. Fearing, ``Towards a soft fingertip with integrated sensing and actuation,'' in \emph{2018 {IEEE/RSJ} International Conference on Intelligent Robots and Systems, {IROS} 2018, Madrid, Spain, October 1-5, 2018}.\hskip 1em plus 0.5em minus 0.4em\relax {IEEE}, 2018.

\bibitem{9197050}
X.~Lin, L.~Willemet, A.~Bailleul, and M.~Wiertlewski, ``Curvature sensing with a spherical tactile sensor using the color-interference of a marker array,'' in \emph{2020 IEEE International Conference on Robotics and Automation (ICRA)}, 2020.

\bibitem{6943124}
C.~Ciliberto, L.~Fiorio, M.~Maggiali, L.~Natale, L.~Rosasco, G.~Metta, G.~Sandini, and F.~Nori, ``Exploiting global force torque measurements for local compliance estimation in tactile arrays,'' in \emph{2014 IEEE/RSJ International Conference on Intelligent Robots and Systems}, 2014.

\bibitem{useOfTactileFeedback}
Z.~Su, J.~Fishel, T.~Yamamoto, and G.~Loeb, ``Use of tactile feedback to control exploratory movements to characterize object compliance,'' \emph{Frontiers in neurorobotics}, vol.~6, p.~7, 07 2012.

\bibitem{Gent1958OnTR}
A.~N. Gent, ``On the relation between indentation hardness and young's modulus,'' \emph{Rubber Chemistry and Technology}, vol.~31, 1958.

\bibitem{He2023}
L.~He, L.~Tao, Z.~Ma, X.~Du, and W.~Wan, ``Cherry tomato firmness detection and prediction using a vision-based tactile sensor,'' \emph{Journal of Food Measurement and Characterization}, vol.~18, no.~2, p. 1053–1064, Nov. 2023.

\bibitem{ERUKAINURE2022107289}
F.~E. Erukainure, V.~Parque, M.~Hassan, and A.~M. FathEl-Bab, ``Estimating the stiffness of kiwifruit based on the fusion of instantaneous tactile sensor data and machine learning schemes,'' \emph{Computers and Electronics in Agriculture}, vol. 201, p. 107289, 2022.

\bibitem{avocadomaturity}
I.~Fahmy, I.~Hussain, N.~Werghi, T.~Hassan, and L.~Seneviratne, ``Hapticformers: Utilizing transformers for avocado maturity grading through vision-based tactile assessment,'' 02 2024.

\bibitem{zeng2019tossingbot}
A.~Zeng, S.~Song, J.~Lee, A.~Rodriguez, and T.~Funkhouser, ``Tossingbot: Learning to throw arbitrary objects with residual physics,'' 2019.

\bibitem{huang2022understanding}
H.-J. Huang, X.~Guo, and W.~Yuan, ``Understanding dynamic tactile sensing for liquid property estimation,'' 2022.

\bibitem{kim2023simultaneoustactileestimationcontrol}
\BIBentryALTinterwordspacing
S.~Kim, D.~K. Jha, D.~Romeres, P.~Patre, and A.~Rodriguez, ``Simultaneous tactile estimation and control of extrinsic contact,'' 2023. [Online]. Available: \url{https://arxiv.org/abs/2303.03385}
\BIBentrySTDinterwordspacing

\bibitem{mahler2017dexnet}
J.~Mahler, J.~Liang, S.~Niyaz, M.~Laskey, R.~Doan, X.~Liu, J.~A. Ojea, and K.~Goldberg, ``Dex-net 2.0: Deep learning to plan robust grasps with synthetic point clouds and analytic grasp metrics,'' 2017.

\bibitem{hibbeler1994mechanics}
R.~C. Hibbeler, \emph{Mechanics of materials}.\hskip 1em plus 0.5em minus 0.4em\relax MacMillan Publishing Company, 1994.

\bibitem{popov2019handbook}
V.~L. Popov, M.~He{\ss}, and E.~Willert, \emph{Handbook of contact mechanics: exact solutions of axisymmetric contact problems}.\hskip 1em plus 0.5em minus 0.4em\relax Springer Nature, 2019.

\bibitem{MDRarticle}
V.~Popov and M.~Heß, ``Method of dimensionality reduction in contact mechanics and friction: A users handbook. i. axially-symmetric contacts,'' \emph{Acta Universitas}, vol.~12, pp. 1--14, 04 2014.

\bibitem{Greaves2011-tp}
G.~N. Greaves, A.~L. Greer, R.~S. Lakes, and T.~Rouxel, ``\BIBforeignlanguage{en}{Poisson's ratio and modern materials},'' \emph{\BIBforeignlanguage{en}{Nat. Mater.}}, vol.~10, no.~11, Oct. 2011.

\bibitem{Calli2017-sw}
B.~Calli, A.~Singh, J.~Bruce, A.~Walsman, K.~Konolige, S.~Srinivasa, P.~Abbeel, and A.~M. Dollar, ``\BIBforeignlanguage{en}{{Yale-CMU-Berkeley} dataset for robotic manipulation research},'' \emph{\BIBforeignlanguage{en}{Int. J. Rob. Res.}}, vol.~36, no.~3, Mar. 2017.

\bibitem{MatWeb}
\BIBentryALTinterwordspacing
``Material property data,'' \emph{MatWeb}, 2024. [Online]. Available: \url{https://www.matweb.com/}
\BIBentrySTDinterwordspacing

\bibitem{10.5254/1.3547752}
H.~J. Qi, K.~Joyce, and M.~C. Boyce, ``Durometer hardness and the stress-strain behavior of elastomeric materials,'' \emph{Rubber Chemistry and Technology}, vol.~76, no.~2, pp. 419--435, 05 2003.

\bibitem{siliconekent}
K.~Larson, ``Can you estimate modulus from durometer hardness for silicones? yes, but only roughly … and you must choose your modulus carefully!'' \emph{Dow White Paper}, 09 2017.

\bibitem{willemet2025physicsinformed}
\BIBentryALTinterwordspacing
L.~Willemet, M.~Burgess, and E.~Adelson, ``Physics-informed online estimation of compliance with gelsight sensor,'' in \emph{8th IEEE-RAS International Conference on Soft Robotics Extended Abstracts}, 2025. [Online]. Available: \url{https://github.com/lwillemet/YoungModulusEstimation/blob/main/Physics-informed_2025_Robosoft.pdf}
\BIBentrySTDinterwordspacing

\end{thebibliography}


\appendix

\subsection{Method of Dimensionality Reduction (MDR)}\label{appendix:MDR}

\begin{figure}[htbp]
        \vspace{-0.25cm}
        \centerline{\includegraphics[width=\linewidth]{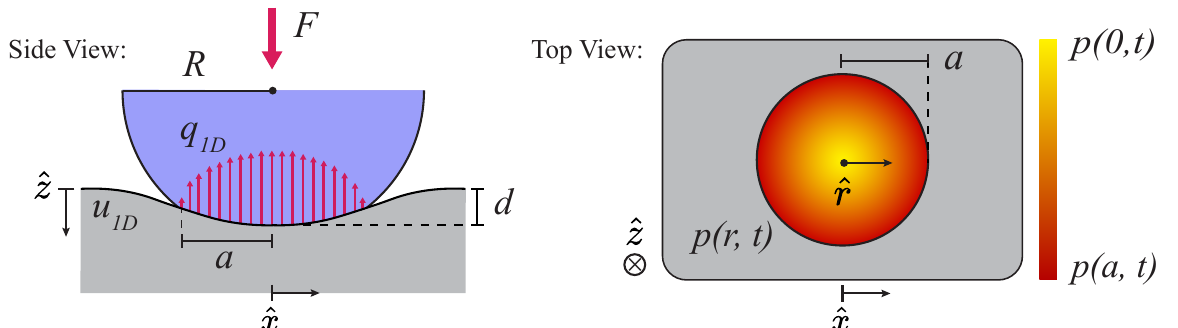}}
        \vspace{-0.25cm}
        \caption[font=small,labelfont=bf]{\textbf{Method of dimensionality reduction (MDR) diagram.} An elastic sphere is colliding with an elasic half-space at time $t$. The 2D radial stress field $p$ is transformed into a 1D force field $q_\text{1D}$.}
        \label{fig:MDR}
        \vspace{-0.25cm}
    \end{figure}

    As discussed in Section~\ref{sec:contactmodels}, we model contact between an unknown object and our flat, camera-based tactile sensor as a collision between an elastic sphere and an elastic half-space using Hertzian contact theory. A mechanical diagram of contact at time $t$ with labeled variables is provided in Fig.~\ref{fig:MDR}.
        
    We compute the maximum pressure of contact $p_0$ between both bodies in Eq.~\ref{eq:p0}. This is a simple expression of force $F$ and contact radius $a$. We can substitute $a$ to create an expression dependent on the aggregate elastic compliance of the bodies $E^{*}$~\cite{johnson1987contact}.
    \begin{equation}
        p_0(t) = \frac{3 F(t)}{2 \pi a^2(t)} = \frac{1}{\pi} {\left(\frac{F(t) {E^*}^2}{R^2}\right)}^{\frac{1}{3}}
        \label{eq:p0}
    \end{equation}
    Using this maximum pressure $p_0$ at timestep $t$, we create the radial pressure field of contact $p(r, t)$. This expression comes directly from Hertzian contact theory~\cite{johnson1987contact}.
    \begin{equation}
        p(r, t) = p_0(t) \sqrt{1 - \frac{r^2}{a^2(t)}}
        \label{eq:pr}
    \end{equation}
    Now, we apply the method of dimensionality reduction (MDR). This technique is used to solve the stress field for surface deformation with the assumption that contact geometries are axisymmetric. By utilizing this method, we can reconstruct the deformation of the sensor surface. We begin by transforming coordinates of the pressure field into 1D force density $q_\text{1D}$ through Eq.~\ref{eq:dim_transform}.
    \begin{equation}
        q_\text{1D}(x, t) = 2 \int_x^\infty{\frac{r p(r, t)}{\sqrt{r^2 - x^2}} dr}
        \label{eq:dim_transform}
    \end{equation}
    From here, we apply a constitutive relationship on the sensor gel pad to get a one-dimensional displacement of the gel surface $u_\text{1D}$ in the normal direction $\hat{z}$. By transforming the stress field to a single dimension, we have simplified the displacement geometry to be axisymmetric about $\hat{z}$ from the center $x = 0$.
    \begin{equation}
        u_\text{1D}(x, t) = \left(\frac{1 - \nu_\text{sensor}^2}{E_\text{sensor}}\right) q_\text{1D}(x, t)
        \label{eq:omega_def}
    \end{equation}
    The maximum displacement will occur at $x = 0$. We define this to be equivalent to our maximum depth $d$.        
    \begin{equation}
        u_\text{1D}(0, t) = d(t) = \left(\frac{1 - \nu_\text{sensor}^2}{E_\text{sensor}}\right) q_\text{1D}(0, t)
        \label{eq:omega_d}
    \end{equation}
    \begin{equation}
        u_\text{1D}(0, t) = d(t) = 2 \left(\frac{1 - \nu_\text{sensor}^2}{E_\text{sensor}}\right) \int_0^{a(t)}{p(r, t) dr}
        \label{eq:simplify_MDR}
    \end{equation}
    We solve this integral to get Eq.~\ref{eq:final_MDR}, which expresses maximum observed surface displacement $d$ in terms of normal force $F$.
    \footnotesize\begin{equation}
        \int_0^{a(t)}{p_0(t) \sqrt{1 - \frac{r^2}{a^2(t)}} dr} = \frac{\pi}{4} p_0(t) a(t) = {\left(\frac{3 F(t) {E^*}^2}{32 R^2}\right)}^{\frac{1}{3}} a(t)
    \end{equation}\normalsize
    \begin{equation}
        u_\text{1D}(0, t) = d(t) = \frac{1 - \nu_\text{sensor}^2}{E_\text{sensor}} {\left(\frac{3 {E^*}^{2} F(t)}{32 R^2(t)}\right)}^{\frac{1}{3}} a(t)
        \label{eq:final_MDR}
    \end{equation}
    We apply this relation for every timestep $t$ over the duration of a grasp. Using measurements of $d$, $a$, and $F$, we can apply a linear least-squares algorithm to retrieve a best fit value $\hat{E}^{*}$. The aggregate modulus $\hat{E}^{*}$ is a function of the grasped object's Young's modulus and known mechanical properties of our sensor's silicone gel pads, as defined in Eq.~\ref{eq:E_star}. Thus, this estimate is used to compute the Young's modulus of the unknown grasped object $\hat{E}_\text{hertz}$, as formulated in Eq.~\ref{eq:E_relation}.


\end{document}